\documentclass[10pt,twocolumn,letterpaper]{article}

\usepackage{iccv}              
\usepackage{float}

\definecolor{iccvblue}{rgb}{0.21,0.49,0.74}
\usepackage[pagebackref,breaklinks,colorlinks,allcolors=iccvblue]{hyperref}
\usepackage{colortbl}

\def\paperID{780} 
\def\confName{ICCV}
\def\confYear{2025}

\title{SA-Occ: Satellite-Assisted 3D Occupancy Prediction in Real World} 

\author{Chen Chen\textsuperscript{\rm 1,\rm 2*}, Zhirui Wang\textsuperscript{\rm 1}\thanks{Co-first authors.} , Taowei Sheng\textsuperscript{\rm 1,\rm 2}, Yi Jiang\textsuperscript{\rm 1,\rm 2}, Yundu Li\textsuperscript{\rm 1,\rm 2}, Peirui Cheng\textsuperscript{\rm 1},\\
Luning Zhang\textsuperscript{\rm 1}, Kaiqiang Chen\textsuperscript{\rm 1}\thanks{Corresponding author.} , Yanfeng Hu\textsuperscript{\rm 1}, Xue Yang\textsuperscript{\rm 3}, Xian Sun\textsuperscript{\rm 1}\\
\vspace{-0.3em}
\small\textsuperscript{\rm 1}Key Laboratory of Target Cognition and Application Technology,\\
\vspace{-0.3em}
\small Aerospace Information Research Institute, Chinese Academy of Sciences\\
\small\textsuperscript{\rm 2}University of Chinese Academy of Sciences \quad \textsuperscript{\rm 3}Shanghai Jiao Tong University\\
\vspace{-0.3em}
{\tt\small chenchen235@mails.ucas.ac.cn}\\
{\tt\small \url{https://github.com/chenchen235/SA-Occ}}
}

\begin{document}
\definecolor{myblue}{RGB}{0,112,192}
\definecolor{myyellow}{RGB}{215,215,0}
\definecolor{myorange}{RGB}{249,104,13}

\maketitle

\begin{abstract}
Existing vision-based 3D occupancy prediction methods are inherently limited in accuracy due to their exclusive reliance on street-view imagery, neglecting the potential benefits of incorporating satellite views. 
We propose \textbf{SA-Occ}, \textbf{the first} Satellite-Assisted 3D occupancy prediction model, which leverages GPS \& IMU to integrate historical yet readily available satellite imagery into real-time applications, effectively mitigating limitations of ego-vehicle perceptions, involving occlusions and degraded performance in distant regions. 
To address the core challenges of cross-view perception, we propose: 
1) \textbf{Dynamic-Decoupling Fusion}, which resolves inconsistencies in dynamic regions caused by the temporal asynchrony between satellite and street views;
 2) \textbf{3D-Proj Guidance}, a module that enhances 3D feature extraction from inherently 2D satellite imagery;
 and 3) \textbf{Uniform Sampling Alignment}, which aligns the sampling density between street and satellite views.
Evaluated on Occ3D-nuScenes, SA-Occ achieves state-of-the-art performance, 
especially among single-frame methods, with a 39.05\% mIoU (a 6.97\% improvement), while incurring only 6.93 ms of additional latency per frame. 
Our code and newly curated dataset are available at \href{https://github.com/chenchen235/SA-Occ}{https://github.com/chenchen235/SA-Occ}.

\end{abstract}
\section{Introduction}
\label{sec:intro}

\begin{figure*}[!t]
    \centering
    \begin{subfigure}[b]{0.24\textwidth}
        \centering
        \includegraphics[width=0.78\textwidth]{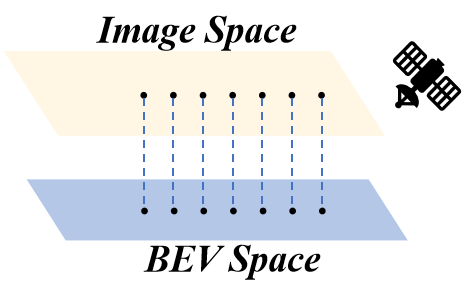}
        \caption{Satellites view: Naturally aligned.} 
        \label{fig:sub1}
    \end{subfigure}
    \hfill 
    \begin{subfigure}[b]{0.375\textwidth}
        \centering
        \includegraphics[width=0.68\textwidth]{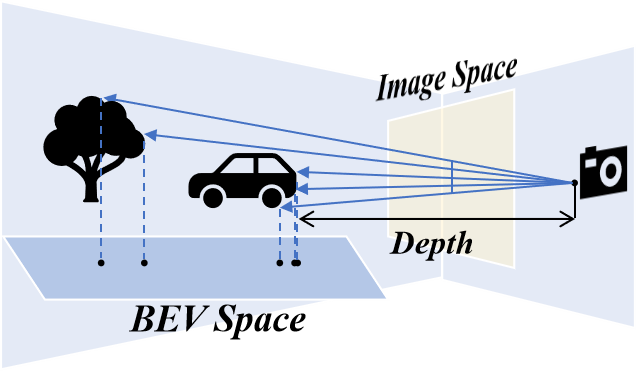}
        \caption{Street view: Depth-based forward projection.}
        \label{fig:sub2}
    \end{subfigure}
    \hfill 
    \begin{subfigure}[b]{0.375\textwidth}
        \centering
        \includegraphics[width=0.68\textwidth]{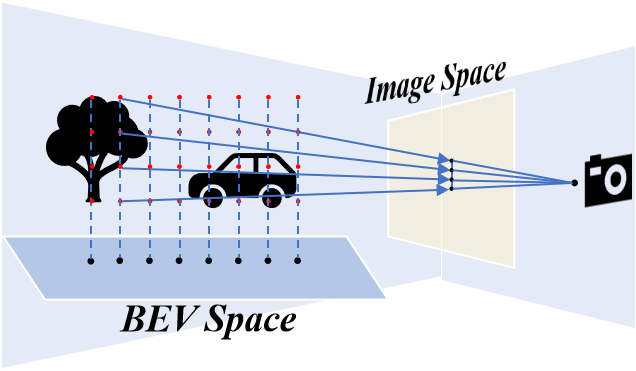}
        \caption{Supplement: Preset-point backward projection.}
        \label{fig:sub3}
    \end{subfigure}
    \caption{
   Comparison of BEV feature acquisition methods: satellite (a: natural alignment with BEV space) vs. street views (b: misalignment due to dense-near-sparse-far characteristic), with our supplement (c: preemptive alignment via predefined points.). 
    }
    \label{fig:Sat_vs_Str}
\end{figure*}


Vision-based 3D occupancy prediction is widely favored in autonomous driving for its cost-effectiveness \cite{zhang2024vision}, yet it often struggles with accuracy limitations \cite{liang2022bevfusion,min2024autonomous}. Existing methods either improve accuracy through increasingly complex models \cite{huang2023tri, wei2023surroundocc, li2023fb, zhang2023occformer}, or seek a balance between accuracy and speed using sparse and lightweight methods \cite{li2023voxformer,yu2023flashocc,hou2024fastocc, lu2023octreeocc, tang2024sparseocc}. 
Despite these advances, all these methods rely exclusively on street views and ignore the benefits of integrating satellite views, 
thereby susceptible to performance degradation caused by occlusions and sparse sampling in distance regions, as shown in Figure \ref{fig:challenge}.


Satellite-street cross-view perception has evolved from coarse-grained retrieval \cite{hu2018cvm} to fine-grained localization \cite{li2023voxformer, lu2023octreeocc, tang2024sparseocc, ye2024cross}, and has further expanded to other tasks such as fine-grained building attribute segmentation \cite{ye2024sg} and cross-view synthesis \cite{shi2022geometry, lin2024geometry, li2024crossviewdiff, ye2024skydiffusion}. These advancements demonstrate the feasibility and potential of satellite-street view collaboration. 
Recent works \cite{gao2024complementing, zhao2024opensatmap} have begun to explore the integration of satellite imagery for improving map construction. However, their approaches rely on the manual alignment of landmark keypoints \cite{gao2024complementing, zhao2024opensatmap}, a process that is largely infeasible for autonomous driving applications. More importantly, their focus remains on static targets within the scope of 2D tasks \cite{gao2024complementing, zhao2024opensatmap}, while neglecting the substantial potential of satellite imagery for applications involving dynamic targets and 3D modeling. 


In this work, unlike manual alignment via landmark keypoints \cite{gao2024complementing, zhao2024opensatmap}, we utilize GPS \& IMU data to obtain the ego-vehicle's pose and extract corresponding satellite image slices from Google Earth \cite{lisle2006google} based on its real-time position and driving direction, 
enabling rapid access to readily available satellite imagery. 
This method creates a novel supplemental dataset complementing OCC3D-Nuscenes \cite{tian2024occ3d}.

Although satellite views provide complementary perception for occluded and long-range areas, new challenges are introduced when considering dynamic targets (see Figure \ref{fig:challenge}). 
The temporal asynchrony between satellite and street views is inevitable, as satellite imagery is pre-existing and captured before street-view observations. This discrepancy leads to inconsistencies in representing dynamic targets, such as moving vehicles and pedestrians, which may be present in one view but absent in the other. 
To address this, we propose a Dynamic-Decoupling Fusion module through a hierarchical dynamic-static attention mechanism. It leverages BEV features from the street view to construct a real-time dynamic region attention weight map, guided by a motion-aware binary decoupling criterion based on target mobility attributes and explicitly supervised through 2D binary semantic maps projected from 3D annotations. During cross-view fusion, the weight map functions as a spatial gate, applying soft masking to satellite features corresponding to dynamic regions in the street view. This establishes a 
\textit{static-dominant, dynamic-suppressed} 
fusion paradigm, effectively reducing interference caused by temporal misalignment of dynamic targets. 

For 3D modeling specifics, satellites and street views have distinct methods for obtaining BEV features, as shown in Figure \ref{fig:Sat_vs_Str}. Satellites offer a top-down perspective, allowing direct extraction of BEV features from imagery (Figure \ref{fig:sub1}). In contrast, street views require complex view transformation due to their ground-level perspective and occlusions (Figure \ref{fig:sub2}). 
This leads to two gaps. First, the implicit 2D feature extractor in the satellite branch struggles to capture 3D information, unlike the explicit 3D-guided view transformation in street views. To address this, we introduce a 3D-Proj Guidance module in the satellite branch, which incorporates semantic supervision for 3D semantic information and height supervision for 3D geometry by projecting 3D occupancy labels into the satellite view. 
Second, the perspective nature of street views results in dense sampling in nearby regions but sparse in distant areas, whereas satellite views provide uniform sampling in BEV space. This disparity limits effective cross-view feature interaction between satellite and street views, particularly in long-range regions. 
To address this, we introduce a Uniform Sampling Alignment module via backward projection (see Figure \ref{fig:sub3}) as a supplement to the precedent forward projection \cite{philion2020lift, li2023bevdepth, yu2023flashocc}, inspired by the dual view transformation for enhancing street-view-only perception \cite{li2024dualbev, li2023fb}. Specifically, we predefine a set of uniformly distributed 3D points in 3D space and project them to obtain the corresponding 2D features. This approach complements the traditional depth-based paradigm, aligning it with the sampling characteristics of satellite views.


 

Our contributions can be summarized as follows:
\begin{itemize}
\item We present the first satellite-assisted 3D occupancy prediction model (SA-Occ), and curated the first dataset that incorporates satellite imagery in real time using pose information from GPS \& IMU.
\item For dynamic inconsistencies, we propose a Dynamic-Decoupling Fusion module to eliminate the uncertainty caused by the temporal asynchrony of satellite images.
\item For 3D modeling specifics, we introduce a 3D-proj Guidance in the satellite branch to bridge the gap between 2D feature extractors and view transformation, and a Uniform Sampling Alignment module in the street branch to bridge the sampling density gap of BEV space.  
\item Our method achieves state-of-the-art performance on Occ3D-NuScenes. Our single-frame model even surpasses the dual-frame models and shows competitive performance compared to models using 8 or 16 frames.
\end{itemize}

\section{Related Work}
\label{sec:related}

\textbf{Vision-Based 3D Occupancy Prediction}: 
MonoScene \cite{cao2022monoscene} pioneered monocular occupancy prediction using a 2D-3D UNet with visual projection, though limited by monocular input constraints. TPVFormer \cite{huang2023tri} advanced the field with tri-view representation, achieving LiDAR-level performance from camera inputs. Subsequent works enhanced accuracy through innovative architectures: SurroundOcc \cite{wei2023surroundocc} adopted a coarse-to-fine strategy with multi-scale depth supervision, OccFormer \cite{zhang2023occformer} introduced a dual-path transformer for joint global-local context modeling, and FBOcc \cite{li2023fb} combined forward projection \cite{philion2020lift} and backward projection \cite{li2024bevformer}  for more complete BEV representations. However, these methods face computational intensity challenges, limiting real-time applicability. 
To address this, sparsification strategies emerged: VoxFormer \cite{li2023voxformer} introduced a sparse voxel transformer, OctreeOcc \cite{lu2023octreeocc} employed octree representations for key 3D regions, and SparseOcc \cite{tang2024sparseocc} introduced a sparse voxel decoder with mask transformer \cite{cheng2022masked} to achieve fully sparse. While these methods reduced resource consumption, they remained constrained by voxel representations and 3D convolutions. Recent advancements like FastOcc \cite{hou2024fastocc} and FlashOcc \cite{yu2023flashocc} overcame this limitation by replacing 3D convolutions with 2D counterparts in the occupancy prediction head, significantly improving efficiency. 
Building on FlashOcc, we introduce the satellite view for the first time, significantly enhancing performance while keeping the model lightweight.

\noindent\textbf{Satellite-Street Cross View Perception}:
Early works like CVM-Net \cite{hu2018cvm} established the foundation for street-to-aerial image matching in coarse localization. Subsequent advancements progressed from coarse-grained retrieval to fine-grained localization: TBL \cite{shi2022accurate} introduced geometric alignment for 3-DoF camera pose estimation, CCVPE \cite{xia2023convolutional} developed an efficient convolutional end-to-end framework, and HC-Net \cite{wang2024fine} employed homography-aware correlation for dense feature matching. The field further evolved with EP-BEV \cite{ye2024cross}, which unified perspectives through BEV representation, and SG-BEV \cite{ye2024sg}, enabling fine-grained building attribute segmentation via satellite-guided street-view feature mapping. Parallel developments in cross-view synthesis \cite{shi2022geometry, lin2024geometry, li2024crossviewdiff, ye2024skydiffusion} explored satellite-street image generation for enhanced data augmentation.
Despite demonstrating satellite-street collaboration potential, existing approaches face limitations in autonomous driving applications. Methods like SatForHDMap \cite{gao2024complementing} and OpenSatMap \cite{zhao2024opensatmap} focus on static map construction, neglecting dynamic object perception and 3D modeling that are crucial for real-world driving. Furthermore, their reliance on manual landmark keypoint selection for cross-view alignment overlooks readily available onboard GPS and IMU data, limiting real-time applicability.
Our work addresses these limitations through three key innovations: 1) leveraging GPS and IMU for automatic cross-view alignment, eliminating manual intervention, 2) introducing the first satellite-assisted 3D occupancy prediction framework capable of handling dynamic object perception challenges arising from non-real-time satellite imagery, and 3) resolving the 3D modeling specific gap in 3D explicit guidance and sampling density of BEV feature caused by viewpoints.


\begin{figure*}
    \centering
    \includegraphics[width=0.95\linewidth]{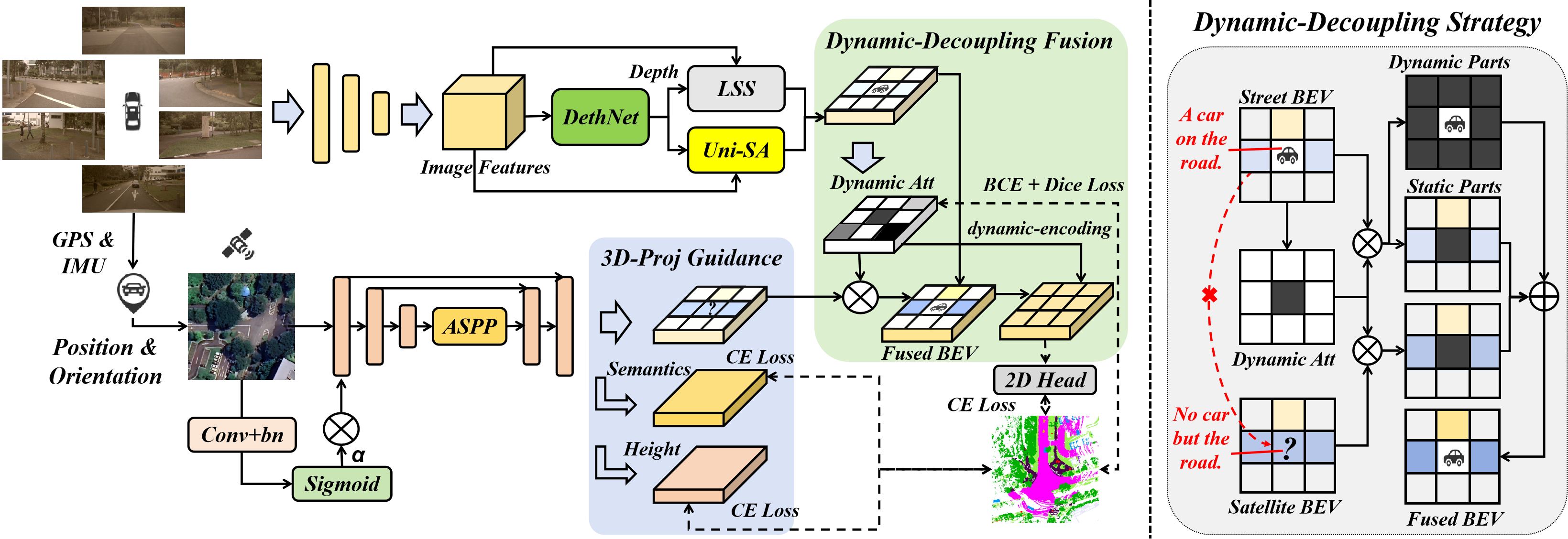}
    \caption{SA-Occ enhances perception with satellite view via GPS \& IMU. It extends the street view branch with a Uniform Sampling Alignment (Uni-SA) module and creates a satellite BEV branch containing a U-shape Feature Extractor with 3D-Proj Guidance module and soft gating. The Dynamic Decoupling Fusion module follows, mitigating satellite interference in dynamic regions via obtaining dynamic region attention from street view with supervision and enhancing dynamic-static region interactions via dynamic-encoding spatial attention.
    }
    \label{fig:method}  
\end{figure*}

\begin{table}[!t]
\fontsize{8.5}{10.5}\selectfont
\centering
\begin{tabular}{ccc}
\toprule
 \textbf{Dataset} & SatforHDMap \cite{gao2024complementing} & Ours\\
\midrule
\textbf{Range} & 30m $\times$ 60m & 80m $\times$ 80m\\
\textbf{Resolution} & (102$\times$202) / (137$\times$273)& (400$\times$400) \\
\textbf{Acquisition Method} & Keypoint Alignment & GPS \& IMU \\
\textbf{Operational Mode} & Non-real-time & Real-time\\
\bottomrule
\end{tabular}
\caption{\label{tab:2}Comparison of our dataset with SatforHDMap.} 
\vspace{-10pt}
\end{table}

\section{Occ3D-NuScenes Extension Dataset}
\label{sec:Complementary}

Occ3D-NuScenes \cite{tian2024occ3d} is a large-scale benchmark for 3D occupancy prediction based on NuScenes \cite{caesar2020nuscenes}, featuring high-quality multi-modal data and covering both dynamic and static objects for comprehensive environment perception.
SatForHDMap \cite{gao2024complementing} serves as a supplementary dataset to NuScenes \cite{caesar2020nuscenes}, focusing on integrating satellite imagery with street-view data for map construction. 
However, it relies on manual keypoint alignment, neglects real-time needs as well as onboard GPS \& IMU data, and its limited perception range is inconsistent with Occ3D-NuScenes \cite{tian2024occ3d}.

The comparison between our dataset and SatForHDMap \cite{gao2024complementing} in Table \ref{tab:2} highlights these limitations. Our work addresses these issues by leveraging the ego vehicle’s GPS \& IMU data to crop satellite images from Google Earth \cite{lisle2006google} directionally, ensuring real-time alignment with the required spatial range and orientation for ego-vehicle perception. Additionally, our dataset offers higher resolution and broader coverage, consistent with Occ3D-Nuscenes \cite{tian2024occ3d}, making it suitable for 3D occupancy prediction.

\section{Method}
\label{sec:method}

This section presents our framework, SA-Occ, as illustrated in \Cref{fig:method}. The framework comprises a satellite BEV branch with 3D-Proj Guidance and a soft gated convolution, a street-view BEV branch with a Uniform Sampling Alignment module, a Dynamic-Decoupling Fusion module for two-branch BEV fusion, and the loss function.

\subsection{Satellite-View BEV Branch}
Satellite view inherently possesses a bird's-eye-view, eliminating the need for view transformation. However, to efficiently extract features, we introduce a 3D-Proj Guidance and a soft gated convolution in the satellite branch.

\subsubsection{U-shape Feature Extractor}

Given an input satellite image \( S \in \mathbb{R}^{H \times W \times 3} \), the lightweight satellite BEV branch first extracts multi-scale feature maps through the partial layers of ResNet-18 \cite{he2016identity}. This yields three hierarchical features:  
\( \mathcal{F}_{1/2} \in \mathbb{R}^{H/2 \times W/2 \times 64} \) (1/2 scale),  
\( \mathcal{F}_{1/4} \in \mathbb{R}^{H/4 \times W/4 \times 128} \) (1/4 scale),  
\( \mathcal{F}_{1/8} \in \mathbb{R}^{H/8 \times W/8 \times 256} \) (1/8 scale).  

To efficiently capture global context, we apply an Atrous Spatial Pyramid Pooling (ASPP) module \cite{chen2017deeplab} to \( \mathcal{F}_{1/8} \), producing enriched features \( \mathcal{F}_{\text{ASPP}} \in \mathbb{R}^{H/8 \times W/8 \times 256} \). These features are then upsampled via bilinear interpolation (\( \uparrow_2 \)) and sequentially fused with \( \mathcal{F}_{1/4} \) and \( \mathcal{F}_{1/2} \) through concatenation (\( \oplus \)):  
\begin{align}
\mathcal{F}_{\text{fused}}^{(1)} &= \mathcal{F}_{1/4} \oplus \uparrow_2(\mathcal{F}_{\text{ASPP}}), \\
\mathcal{F}_{\text{fused}}^{\text{satellite}} &= \mathcal{F}_{1/2} \oplus \uparrow_2(\mathcal{F}_{\text{fused}}^{(1)}),
\end{align}
resulting in a final fused feature \( \mathcal{F}_{\text{fused}}^{\text{satellite}} \in \mathbb{R}^{H/2 \times W/2 \times 256} \).

\subsubsection{3D-Projection Guidance}

While the U-shaped architecture effectively integrates multi-scale features with global-local characteristics, 2D feature extractors struggle to obtain effective 3D information without explicit guidance like view transformation \cite{philion2020lift,li2024dualbev,li2023bevdepth} in street view. To address this, we introduce satellite-view semantic and height supervision (height for 3D geometry and semantics for 3D semantic information), both ground truth (\( \mathcal{G}^{sem} \) and \( \mathcal{G}^{h} \)) derived through projection from the 3D occupancy ground truth \( \mathcal{G}^{occ} \in \mathbb{R}^{X \times Y \times Z} \).

To mitigate the temporal inconsistency caused by the non-real-time nature of satellite imagery, we eliminate dynamic object interference through static-focused projection. Let \( \mathcal{C}_{dyn} = \{0,1,...,10\} \) denote dynamic object categories (e.g., cars, pedestrians) and \( \mathcal{C}_{sta} = \{11,...,18\} \) represent static categories (e.g., drivable surface, sidewalks). We first generate a 3D static mask \( \mathcal{M}_{sta} \in \{0,1\}^{X \times Y \times Z} \) by filtering occupancy ground truth:  
\begin{align}
\mathcal{M}_{sta}(i,j,k) = \begin{cases} 
1, & \text{if } \mathcal{G}^{occ}(i,j,k) \in \mathcal{C}_{sta}; \\
0, & \text{otherwise}.
\end{cases}    
\end{align}

We then encode the vertical height information by element-wise multiplying \( \mathcal{M}_{sta} \) with a z-axis height encoding tensor \( \mathcal{E}_z \in \mathbb{R}^{X \times Y \times Z} \), where \( \mathcal{E}_z(i,j,k) = k \). The resulting height-weighted static voxel \( \mathcal{V}_{sta} = \mathcal{M}_{sta} \odot \mathcal{E}_z \) preserves only static categories with their vertical positions.  

The satellite-view height map \( \mathcal{G}^{h} \in \mathbb{R}^{H \times W} \) is derived by selecting the maximum height value per BEV grid:  
\begin{align}
\mathcal{G}_{sat}(i,j) = \arg\max_k \mathcal{V}_{sta}(i,j,k).
\end{align}

Simultaneously, the semantic map \( \mathcal{G}^{sem} \in \mathbb{R}^{H \times W} \) is obtained by indexing the occupancy prediction with \( \mathcal{G}^{h} \):  
\begin{align}
\mathcal{G}^{sem}(i,j) = \mathcal{G}^{occ}\left(i,j,\mathcal{G}^{h}(i,j)\right).
\end{align}


Semantic and height information exhibit stronger correlations compared to the street view due to the top-down nature of satellite imagery \cite{chen2024semstereo}. To leverage this, we maximize feature sharing by processing the fused feature \( \mathcal{F}_{\text{fused}}^{satellite}\) through three parallel \( 3 \times 3 \) convolutional layers, serving as the semantic supervision head, height estimation head, and satellite BEV feature extraction head, respectively. 

\subsubsection{Soft Gated Convolution}

Satellite images often contain high-frequency noise that is inconsistent with street views. 
To mitigate this, we introduce a spatial gated attention module to filter out these noises effectively at the first layer of networks (see Figure \ref{fig:method}), inspired by image inpainting methods \cite{yu2019free}. 
We first construct a convolutional layer \( \mathcal{C} \) with a stride of 2 and Batch Normalization, followed by a sigmoid function \( \sigma \):  
\begin{align}
\alpha = \sigma(\mathcal{C}(S)) \in [0,1]^{H \times W \times C}.
\end{align}  

Then utilize these attention weights to modulate the original features via element-wise multiplication \( \odot \): 
\begin{align}
\mathcal{F}_{1/2}^{\text{filtered}} = \mathcal{F}_{1/2} \odot \alpha.
\end{align}  

This simple module adaptively suppresses high-frequency noise while preserving consistent static features. 

\subsection{Street-view Branch with Uniform Sampling}

While FlashOcc \cite{yu2023flashocc} achieves high-efficiency performance through its 2D prediction head, the sparse BEV features generated by LSS \cite{philion2020lift} limit fusion effectiveness with dense satellite BEV features. To address this, we introduce a Uniform Sampling Alignment (Uni-SA) module, inspired by \cite{li2024dualbev, li2023fb}, that projects pre-defined uniformly distributed 3D point clouds \( \mathcal{P} = \{ (x_i, y_i, z_i) \}_{i=1}^N \) onto multi-view image planes for aligning with satellite uniform dense sampling.  

Formally, given street-view features \( I \in \mathbb{R}^{V \times H \times W \times C} \) (where \( V \) denotes the number of cameras), the LSS branch outputs sparse BEV features \( \mathcal{F}_{\text{LSS}} \in \mathbb{R}^{X \times Y \times D} \), while the Uni-SA generates features \( \mathcal{F}_{\text{Uni-SA}} \in \mathbb{R}^{X \times Y \times C} \) through: 
\begin{align}
\mathcal{F}_{\text{Uni-SA}} = \small\sum_{v=1}^V \text{Interp} \left( \text{Project}(\mathcal{P}, \theta_v), I_v \right),
\end{align}  
where \( \theta_v \) represents camera parameters and \( \text{Interp}(\cdot) \) denotes bilinear interpolation. 

Then, we fuse $\mathcal{F}_{\text{LSS}}$ and $\mathcal{F}_{\text{Uni-SA}}$ to obtain the final street BEV features $\mathcal{F}_{\text{fused}}^{\text{street}}$ via Dual Feature Fusion \cite{li2024dualbev}.

Our method differs from \cite{li2024dualbev} by eliminating the probabilistic correspondences dedicated to foreground extraction in 3D object detection and employing uniform sampling points in 3D space for 3D occupancy prediction.
   
  
\subsection{Dynamic-Decoupling Fusion}  

The direct fusion of satellite and street BEV features leads to inconsistencies in dynamic regions, as satellite imagery cannot capture real-time moving objects. To mitigate this issue, we propose a Dynamic-Decoupling Fusion module.  

The module begins by estimating a dynamic region attention map \(\hat{\mathcal{M}}_{dyn}^{bev}\in \mathbb{R}^{X \times Y} \) from the street-view BEV features \( \mathcal{F}_{\text{fused}}^{\text{street}} \) via Spatial Attention Enhanced ProbNet \cite{li2024dualbev}. Let $F_{dyn}$ be the pre-activation attention features and the dynamic attention map is obtained as $\hat{\mathcal{M}}_{dyn}^{bev} = \sigma(F_{dyn})$, where $\sigma$ denotes the sigmoid function (the same hereafter).

The ground truth \( \mathcal{M}_{dyn}^{bev} \) for explicit supervision is derived by projecting dynamic-class voxels from the 3D occupancy labels \( \mathcal{G}^{\text{occ}} \) and 3D bounding boxes of objects \( \mathcal{G}^{\text{obj}} \):
\begin{equation}
\begin{aligned}
\mathcal{M}_{dyn}^{\text{occ}\rightarrow{\text{bev}}} &= \text{Project}_{3d\rightarrow bev}(\mathcal{G}^{\text{occ}} \odot (1-\mathcal{M}_{sta})),\\
\mathcal{M}_{dyn}^{\text{obj}\rightarrow{\text{bev}}} &= \text{Project}_{3d\rightarrow bev}(\mathcal{G}^{\text{obj}}),\\
\mathcal{M}_{dyn}^{bev} & = \mathcal{M}_{dyn}^{\text{obj}\rightarrow{\text{bev}}} \cup \mathcal{M}_{dyn}^{\text{occ}\rightarrow{\text{bev}}}.
\end{aligned}
\end{equation}
where \( \odot \) represents element-wise multiplication.

To decouple the fusion process, we use the static attention \( (1-\hat{\mathcal{M}}_{dyn}^{bev}) \) as a consistency metric between the satellite and street features. The satellite BEV features \( \mathcal{F}^{\text{sat}}_{\text{fused}} \in \mathbb{R}^{X \times Y \times C} \) are modulated by \( (1-\hat{\mathcal{M}}_{dyn}^{bev}) \) and concatenated with the ground features \( \mathcal{F}_{\text{ground}} \):  
\begin{align}
\mathcal{F}_{\text{fused}} = \text{Conv}\left( \left[ \mathcal{F}^{\text{sat}}_{\text{fused}} \odot (1-\hat{\mathcal{M}}_{dyn}^{bev}) ;\, \mathcal{F}^{\text{ground}}_{\text{fused}} \right] \right),
\end{align}  
where \( [\cdot;\cdot] \) denotes channel-wise concatenation. 


A spatial attention mechanism with dynamic encoding is followed to enhance dynamic perception through adaptive interaction between refined static and dynamic regions while preserving the distinct representation of both regions:
\begin{align}
\mathcal{F}_{\text{final}} = \sigma \left[\text{SA}( \mathcal{F}_{\text{fused}}) + F_{dyn} \right] \odot \mathcal{F}_{\text{fused}},
\end{align}  
where SA denotes spatial attention \cite{woo2018cbam}. 

\subsection{Loss Function}  
We use Cross-Entropy (CE) Loss for satellite semantic segmentation $\mathcal{L}_{\text{sem}}$, satellite height estimation $\mathcal{L}_{\text{hgt}}$, street depth estimation  \( \mathcal{L}_{\text{depth}} \) and 3D occupancy prediction \( \mathcal{L}_{\text{occ}} \).  


For dynamic regions prediction, we augment the CE loss with Dice Loss to address the class imbalance and improve boundary alignment:  
\begin{align}
\mathcal{L}_{\text{dyn}} = \text{CE}(\hat{\mathcal{M}}_{dyn}^{bev}, \mathcal{M}_{dyn}^{bev}) + \text{Dice}(\hat{\mathcal{M}}_{dyn}^{bev}, \mathcal{M}_{dyn}^{bev}).
\end{align}  

The total loss is a weighted sum:  
\begin{equation}
\begin{aligned}
\mathcal{L}_{\text{total}} = &  \lambda_0 \cdot \mathcal{L}_{\text{sem}} + \lambda_1 \cdot \mathcal{L}_{\text{hgt}} \\ + & \lambda_2 \cdot \mathcal{L}_{\text{depth}} + 
\lambda_3 \cdot \mathcal{L}_{\text{dyn}} + \mathcal{L}_{\text{occ}}.
\end{aligned}
\end{equation}

\section{Experiments}
\label{sec:Experiments}



\begin{table*}[!t]
\centering
\fontsize{8.5}{10.5}\selectfont
\setlength{\tabcolsep}{3pt}
\begin{minipage}[t]{0.575\textwidth}
\setlength{\tabcolsep}{1.85pt}
\centering
\begin{tabular}{@{} c|ccc|cccc @{}}
\toprule
\textbf{Model} & \textbf{SAT.} & \textbf{Uni-SA} & \textbf{DDF} & \textbf{mIOU (\%)} & \textbf{D mIOU (\%)} & \textbf{S mIOU (\%)} \\
\hline
Baseline \cite{yu2023flashocc} & &  && 32.08 & 23.38 & 48.02\\
 - & \checkmark & & & 36.89 (+ 4.81) & 27.40 (+ 4.02) & 54.30 (+ 6.28)\\
- & \checkmark & & \checkmark& 37.16(+ 5.08) & 27.93 (+ 4.45) & 54.07 (+ 6.05)\\
- & \checkmark & \checkmark && 38.55 (+ 6.47)& 29.62 (+ 6.24) & \textbf{54.93 (+ 6.91)}\\
SA-Occ (V1) & \checkmark & \checkmark & \checkmark & \textbf{39.05 (+ 6.97)} & \textbf{30.59 (+ 7.21)}& 54.55 (+ 6.53)\\
\bottomrule
\end{tabular}
\caption{Ablation of main structural components. \textbf{SAT.}: overall satellite branch, \textbf{Uni-SA}: Uniform Sampling Alignment, \textbf{DDF}: Dynamic-Decoupling Fusion.}
\label{tab:ablation_structural}
\end{minipage}
\hfill
\begin{minipage}[t]{0.41\textwidth}
\setlength{\tabcolsep}{1.4pt}
\centering
\begin{tabular}{@{} c|ccccccccc @{}}
\toprule
\textbf{Strategy} & \textbf{mIOU (\%)} & \textbf{D mIOU (\%)} & \textbf{S mIOU (\%)} \\
\hline
MaskSegAtt \cite{gao2024complementing} & 23.23 & 8.51 & 50.47\\
Add & 38.57 & 29.70 & 54.84\\
Concat & 38.55 & 29.62 & \textbf{54.93}\\
DDF + Align & 38.78 & 30.13 & 54.63\\
DDF & \textbf{39.05} & \textbf{30.59}& 54.55\\
\bottomrule
\end{tabular}
\caption{Quantitative comparison between our Dynamic-Decoupling Fusion (DDF) and other fusion methods.}
\label{tab:ablation_fusion}
\end{minipage}

\vspace{1em}

\begin{minipage}[t]{0.49\textwidth}
\centering
\begin{tabular}{@{} ccc|cccccc @{}}
\toprule
\textbf{Sat} & \textbf{3D-Proj} & \textbf{Gate} & \textbf{mIOU (\%)} & \textbf{D mIOU (\%)} & \textbf{S mIOU (\%)} \\
\hline
 & & & 32.08 & 23.38 & 48.02\\
\checkmark & & &36.6 (+ 4.52) & \textbf{27.48 (+ 4.1)} & 53.22 (+ 5.2)\\
\checkmark & \checkmark& & 36.82 (+ 4.74)& 27.38 (+ 4.0) & 54.13 (+ 6.11)\\
\checkmark & \checkmark& \checkmark& \textbf{36.89 (+ 4.81)} & 27.40 (+ 4.02) & \textbf{54.30 (+ 6.28)}\\
\bottomrule
\end{tabular}
\caption{Ablation of satellite branch. \textbf{Sat.}: the barely satellite branch, \textbf{3D-Proj}: 3D-Proj Guidance, \textbf{Gate}: Soft Gate mechanism.}
\label{tab:ablation_satellite}
\end{minipage}
\hfill
\begin{minipage}[t]{0.49\textwidth}
\centering
\begin{tabular}{@{} ccc|cccc @{}}
\toprule
\textbf{DDS} & \textbf{SA} & \textbf{DSA} & \textbf{mIOU (\%)} & \textbf{D mIOU (\%}) & \textbf{S mIOU (\%)} \\
\hline
& & & 38.55 & 29.62 & \textbf{54.93}\\
\checkmark & & & 38.88 (+ 0.33)& 30.27 (+ 0.65) & 54.64\\
\checkmark & \checkmark & & 38.95 (+ 0.40)& 30.43 (+ 0.81) & 54.56\\
\checkmark &  & \checkmark & \textbf{39.05 (+ 0.50)} & \textbf{30.59 (+ 0.97)} & 54.55\\
\bottomrule
\end{tabular}
\caption{Ablation of DDF. \textbf{DDS}: Dynamic-Decoupling strategy, \textbf{SA}: Spatial Attention, \textbf{DSA}: Spatial Attention with Dynamic-encoding.}
\label{tab:ablation_ddf}
\end{minipage}

\vspace{1em}

\begin{minipage}[t]{0.42\textwidth}
\setlength{\tabcolsep}{1.2pt}
\centering
\begin{tabular}{@{} c|c|ccc @{}}
\toprule
&\textbf{Source} & \textbf{mIOU (\%)} & \textbf{D mIOU (\%)} & \textbf{S mIOU (\%)} \\
\hline
1& 3d occupancy & 37.69 & 30.01 & 51.79\\
2&3d object box & 37.81 (+ 0.12) & 30.21 (+ 0.2) & 51.74 (- 0.05)\\
3 & $1 \cup 2$ & \textbf{37.94 (+ 0.25)} & \textbf{30.29 (+ 0.28)}& \textbf{51.98 (+ 0.19)}\\
\bottomrule
\end{tabular}
\caption{Ablation of dynamic area labels source.}
\label{tab:ablation_labels}
\end{minipage}
\hfill
\begin{minipage}[t]{0.56\textwidth}
\setlength{\tabcolsep}{1.2pt}
\centering
\begin{tabular}{@{} c|cccc @{}}
\toprule
\textbf{Model} & \textbf{FLOPs (G)} & \textbf{Params (M)} & \textbf{Latency (ms)} & \textbf{mIOU (\%)} \\
\hline
BEVDetOcc \cite{huang2021bevdet} & \textbf{241.76} & \textbf{29.02} & 149.19 & 31.64\\
FlashOcc (M1) \cite{yu2023flashocc} & 248.57 & 44.74 & \textbf{80.65 (-68.54)} & 32.08 (+ 0.44)\\
SA-Occ (V1) & 442.19 & 86.43 & 87.58 (-61.61) & \textbf{39.05 (+ 7.41)}\\
\bottomrule
\end{tabular}
\caption{Quantitative comparison with baselines in terms of accuracy, latency, and model complexity. Tested on a single Nvidia A40 GPU.}
\label{tab:model_complexity}
\end{minipage}

\end{table*}

\definecolor{fleshtone}{RGB}{245, 222, 179}
\begin{table*}[!t]
\fontsize{8.5}{10.5}\selectfont
\setlength{\tabcolsep}{1.75pt}
\centering
\begin{tabular}{c|c|c|c|c|ccccccccccccccccccccccc}
\toprule
& \rotatebox{90}{\textbf{Frames}} & \rotatebox{90}{\textbf{Backbone}} & \rotatebox{90}{\textbf{Input Size}} & \rotatebox{90}{\textbf{mIoU}} & \rotatebox{90}{\textcolor{black}{\rule{1ex}{1ex}} \textbf{others}} & \rotatebox{90}{\textcolor{orange}{\rule{1ex}{1ex}} \textbf{barrier}} & \rotatebox{90}{\textcolor{pink}{\rule{1ex}{1ex}} \textbf{bicycle}} & \rotatebox{90}{\textcolor{yellow}{\rule{1ex}{1ex}} \textbf{bus}} & \rotatebox{90}{\textcolor{blue}{\rule{1ex}{1ex}} \textbf{car}} & \rotatebox{90}{\textcolor{cyan}{\rule{1ex}{1ex}} \textbf{cons.veh.}} & \rotatebox{90}{\textcolor{olive}{\rule{1ex}{1ex}} \textbf{motorcycle}} & \rotatebox{90}{\textcolor{red}{\rule{1ex}{1ex}} \textbf{pedestrian}} & \rotatebox{90}{\textcolor{fleshtone}{\rule{1ex}{1ex}} \textbf{traffic cone}} & \rotatebox{90}{\textcolor{brown}{\rule{1ex}{1ex}} \textbf{trailer}} & \rotatebox{90}{\textcolor{violet}{\rule{1ex}{1ex}} \textbf{truck}} & \rotatebox{90}{\textcolor{purple!60!white}{\rule{1ex}{1ex}} \textbf{drive. surf.}} & \rotatebox{90}{\textcolor{gray}{\rule{1ex}{1ex}} \textbf{other flat}} & \rotatebox{90}{\textcolor{darkgray}{\rule{1ex}{1ex}} \textbf{sidewalk}} & \rotatebox{90}{\textcolor{green!20!white}{\rule{1ex}{1ex}} \textbf{terrain}} & \rotatebox{90}{\textcolor{lightgray}{\rule{1ex}{1ex}} \textbf{manmade}} & \rotatebox{90}{\textcolor{green}{\rule{1ex}{1ex}} \textbf{vegetation}} \\
\hline
MonoScene \cite{cao2022monoscene} & 1 & R101 & (600, 928) & 6.06 & 1.8 & 7.2 & 4.3 & 4.9 & 9.4 & 5.7 & 4.0 & 3.0 & 5.9 & 4.5 & 7.2 & 14.9 & 6.3 & 7.9 & 7.4 & 1.0 & 7.7\\
BEVFormer \cite{li2024bevformer} & 1 & R101 & (600, 928) & 26.88 & 5.9 & 37.8 & 17.9 & 40.4 & 42.4 & 7.4 & 23.9 & 21.8 & 21.0 & 22.4 & 30.7 & 55.4 & 28.4 & 36.0 & 28.1 & 20.0 & 17.7\\
TPVFormer \cite{huang2023tri} & 1 & R101 & (600, 928) & 27.83 & 7.2 & 38.9 & 13.7 & 40.8 & 45.9 & 17.2 & 20.0 & 18.9 & 14.3 & 26.7 & 34.2 & 55.7 & 35.5 & 37.6 & 30.7 & 19.4 & 16.8\\
OccFormer \cite{zhang2023occformer} & 1 & R101 & (600, 928) & 21.93 & 5.9 & 30.3 & 12.3 & 34.4 & 39.2 & 14.4 & 16.5 & 17.2 & 9.3 & 13.9 & 26.4 & 51.0 & 31.0 & 34.7 & 22.7 & 6.8 & 7.0\\
CTF-Occ \cite{tian2024occ3d} & 1 & R101 & (600, 928) & 28.53 & 8.1 & 39.3 & 20.6 & 38.3 & 42.2 & 16.9 & 24.5 & 22.7 & 21.1 & 23.0 & 31.1 & 53.3 & 33.8 & 38.0 & 33.2 & 20.8 & 18.0\\
BEVDetOcc \cite{huang2021bevdet} & 1 & R50 & (256, 704) & 31.64 & 6.7 & 37.0 & 8.3 & 38.7 & 44.5 & 15.2 & 13.7 & 16.4 & 15.3 & 27.1 & 31.0 & 78.7 & 36.5 & 48.3 & 51.7 & 36.8 & 32.1\\
FlashOcc (M1) \cite{yu2023flashocc} & 1 & R50 & (256, 704) & 32.08 & 6.7 & 37.7 & 10.3 & 39.6 & 44.4 & 14.9 & 13.4 & 15.8 & 15.4 & 27.4 & 31.7 & 78.8 & 38.0 & 48.7 & 52.5 & 37.9 & 32.2\\
\rowcolor{gray!20} \textbf{SA-Occ (V1) (Ours)} & \textbf{1} & \textbf{R50} & \textbf{(256, 704)} & \textbf{39.05} & \textbf{10.8} & \textbf{45.9} & \textbf{20.5} & \textbf{46.6} & \textbf{51.1} & \textbf{23.0} & \textbf{22.7} & \textbf{23.1} & \textbf{21.4} & \textbf{33.3} & \textbf{38.2} & \textbf{82.6} & \textbf{43.8} & \textbf{54.0} & \textbf{58.5} & \textbf{47.0} & \textbf{41.4}\\
\hline
BEVDetOcc (2f) \cite{huang2021bevdet} & 2 & R50 & (256, 704) & 36.01 & 8.2 & 44.2 & 10.3 & 42.1 & 49.6 & 23.4 & 17.4 & 21.5 & 19.7 & 31.3 & 37.1 & 80.1 & 37.4 & 50.4 & 54.3 & 45.6 & 39.6\\
BEVDetOcc (2f) \cite{huang2021bevdet} & 2 & R50 & (384, 704) & 37.28 & 8.8 & 45.2 & 19.1 & 43.5 & 50.2 & 23.7 & 19.8 & 22.9 & 20.7 & 31.9 & 37.7 & 80.3 & 37.0 & 50.5 & 53.4 & 47.1 & 41.9\\
FlashOcc (M2) \cite{yu2023flashocc} & 2 & R50 & (256, 704) & 37.84 & 9.1 & 46.3 & 17.7 & 42.7 & 50.6 & 23.7 & 20.1 & 22.3 & 24.1 & 30.3 & 37.4 & 81.7 & 40.1 & 52.3 & 56.5 & 47.7 & 40.6 \\
\rowcolor{gray!20} \textbf{SA-Occ (V2) (Ours)} & \textbf{2} & \textbf{R50} & \textbf{(256, 704)} & \textbf{40.65} & \textbf{10.9} & \textbf{48.5} & \textbf{23.5} & \textbf{45.8} &\textbf{52.8} & \textbf{24.5} & \textbf{23.6} & \textbf{24.2} & \textbf{22.7} & \textbf{35.0} & \textbf{40.3} & \textbf{83.5} & \textbf{44.1} & \textbf{55.5} & \textbf{60.0} & \textbf{51.2} & \textbf{45.3}\\
\hline
BEVDetOcc (8f) \cite{huang2021bevdet} & 8 & R50 & (384, 704) & 39.26 & 9.3 & 47.1 & 19.2 & 41.5 & 52.2 & \textbf{27.2} & 21.2 & 23.3 & 21.6 & \textbf{35.8} & 38.9 & 82.5 & 40.4 & 53.8 & 57.7 & 49.9 & 45.8\\
FBOcc (E) \cite{li2023fb} & 16 & R50 & (256, 704) & 39.11 & \textbf{13.6} & 44.7 & \textbf{27.0} & 45.4 & 49.1 & 25.2 & 26.3 & 27.7 & 27.8 & 32.3 & 36.8 & 80.1 & 42.8 & 51.2 & 55.1 & 42.2 & 37.5\\
FastOcc \cite{hou2024fastocc} & 16 & R101 & (640, 1600) & 39.21 & 12.1 & 43.5 & 28.0 & 44.8 & 52.2 & 23.0 & 29.1 & 29.7 & 27.0 & 30.8 & 38.4 & 82.0 & 41.9 & 51.9 & 53.7 & 41.0 & 35.5\\
FastOcc-TTA \cite{hou2024fastocc} & 16 & R101 & (640, 1600)& 40.75 & 12.9 & 46.6 & 29.9 & \textbf{46.1} & 54.1 & 23.7 & \textbf{31.1} & \textbf{30.7} & \textbf{28.5} & 33.1 & 39.7 & 83.3 & 44.7 & 53.9 & 55.5 & 42.6 & 36.5\\ 
\rowcolor{gray!20} \textbf{SA-Occ (V3) (Ours)} & \textbf{8} & \textbf{R50} & \textbf{(256, 704)} & \textbf{41.69} & 11.9 & \textbf{50.5} & 22.1 & 45.0 & \textbf{54.2} & 25.9 & 24.0 & 24.4 & 26.3 & 35.4 & \textbf{41.0} & \textbf{84.3} & \textbf{46.1} & \textbf{57.0} & \textbf{61.2} & \textbf{52.8} & \textbf{46.8}\\
\hline
BEVDetOcc \cite{huang2021bevdet} & 2 & Swin-B & (512, 1408) & 42.02 & 12.2 & 48.6 & 25.1 & 52.0 & 54.5 & 27.9 & 28.0 & 28.9 & 27.2 & 36.4 & 42.2 & 82.3 & 43.3 & 54.6 & 57.9 & 48.6 & 43.6\\
BEVDetOcc* \cite{huang2021bevdet} & 2 & Swin-B & (512, 1408) & 42.45 & 12.4 & 50.2 & 27.0 & 51.9 & 54.7 & \textbf{28.4} & 29.0 & 29.0 & 28.3 & 37.1 & 42.5 & 82.6 & 43.2 & 54.9 & 58.3 & 48.8 & 43.8\\
FlashOcc (M3) \cite{yu2023flashocc} & 2 & Swin-B & (512, 1408) & 43.52 & 13.4 & 51.1 & 27.7 & 51.6 & 56.2 & 27.3 & 30.0 & 29.9 & 29.8 & 37.8 & 43.5 & 83.8 & 46.6 & 56.2 & 59.6 & 50.8 & 44.7\\
\rowcolor{gray!20} SA-Occ (V4) (Ours) & 2 & Swin-B & (512, 1408) & 43.90 & 13.2 & 50.8 & 28.5 & 52.1 & 56.9 & 26.0 & 31.2 & 29.6 & 28.8 & \textbf{38.5} & 44.0 & 84.3 & \textbf{46.6} & 56.9 & 61.4 & 51.7 & 46.0\\
\rowcolor{gray!20} SA-Occ (V5) (Ours) & 2 & Swin-B & (512, 1408) & 44.29 & 13.3 & 51.0 & 28.7 & \textbf{52.7} & 56.8 & 27.3 & 30.4 & 29.7 & 29.4 & 37.5 & 43.5 & 84.8 & 46.5 & 58.3 & 62.1 & 52.8 & 48.3\\
\rowcolor{gray!20} \textbf{SA-Occ (V5)* (Ours)} & \textbf{2} & \textbf{Swin-B} & \textbf{(512, 1408)} & \textbf{44.64} & \textbf{13.7} & \textbf{51.1} & \textbf{30.4} & 52.4 & \textbf{56.9} & 27.0 & \textbf{31.6} & \textbf{30.2} & \textbf{30.3} & 37.6 & \textbf{44.0} & \textbf{84.8} & 46.1 & \textbf{58.6} & \textbf{62.1} & \textbf{53.1} & \textbf{49.1}\\

\bottomrule
\end{tabular}
\caption{\label{tab: 4}3D Occupancy prediction performance on the Occ3D-nuScenes dataset. All other results are obtained from the official declaration. \textbf{*}: 12 epochs extra training compared to normal 24 epochs. \textbf{Bold}: Best.}
\end{table*}

\begin{figure*}
    \centering
    \includegraphics[width=0.95\linewidth]{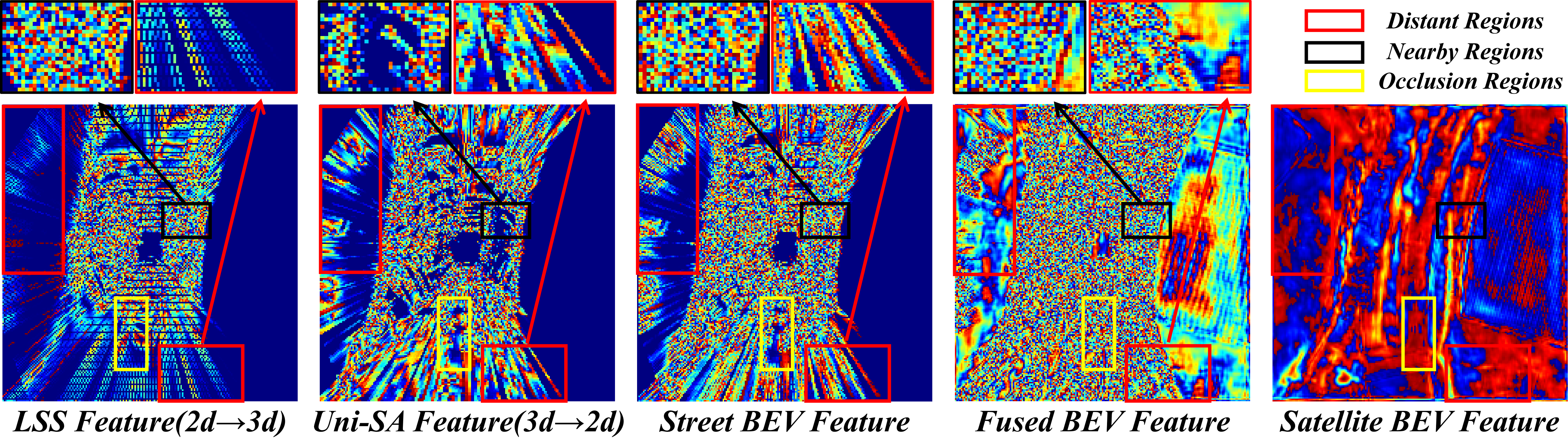}
    \caption{Qualitative comparison of BEV features at different stages. (a) The LSS-based 2D-to-3D forward-proj view transformation generates dense features in nearby regions but suffers from sparsity at distant regions due to its radial projection pattern. By adding 3D-to-2D uniform sampling that aligns with satellite view, we supplement features at distant regions, maximize street feature density, and thereby optimize the interaction and fusion of satellite and street-view features across both nearby and distant regions. (b) Even with maximized feature density, the street-view perspective remains limited by occlusions, which are effectively supplemented by the satellite viewpoint. }
    \label{fig: Qualitative_feature}
\end{figure*}

\subsection{Metrics}
We use mIoU (mean Intersection-over-Union) as the primary metric. To highlight improvements, we measure D-mIoU (Dynamic mIoU) for 11 movable object categories (classes 0-10) and S-mIoU (Static mIoU) for 6 non-movable object categories (classes 11-18).

\subsection{Implementation Details}
We conduct experiments on NVIDIA A40 GPUs. The hyperparameters for the loss function are set as follows: $\lambda_0=0.5$, $\lambda_1=0.05$, $\lambda_2=0.05$, $\lambda_3=0.2$. 
The backbone of our satellite branch is ResNet18 \cite{he2016identity} by default for efficiency. All training configurations align with our baseline \cite{yu2023flashocc}, using the Adam optimizer, a learning rate of \( 2 \times 10^{-4} \), 24 training epochs, a BEV grid resolution of \( 200 \times 200 \), and the CBGS data augmentation method \cite{zhu2019class}.

\subsection{Ablation Study}
\textbf{Impact of Main Structural Components:} We experimentally explore improvements from each main structural component, as shown in Table \ref{tab:ablation_structural}. 
Starting with FlashOcc \cite{yu2023flashocc} as the baseline, we first introduce the complete satellite BEV branch, which results in a +4.81\% mIoU improvement, with the dynamic categories a +4.02\% mIoU gain and a more significant improvement of +6.28\% Static mIoU. This demonstrates the strong complementary effect of the satellite view on street-view perception. The qualitative comparison (Figure \ref{fig: Qualitative_feature}) also reveals that integrating the satellite view brings additional perception of occluded and distant regions, as well as boundary information. Building upon this, we incorporate the Uniform Sampling Alignment (Uni-SA) into the street branch, which yields an additional +1.66\% mIoU improvement, benefiting both dynamic and static object categories. Finally, we integrate the Dynamic-Decoupling Fusion (DDF) module, achieving a further +0.50\% mIoU gain. Notably, the dynamic object categories exhibit a +0.97\% mIoU improvement, resulting in a final cumulative gain for dynamic objects (+7.21\% mIoU) that surpasses that of static (+6.53\% mIoU), as shown in Line 5. This confirms the effectiveness of the DDF module in enhancing the perception of dynamic objects.  
Furthermore, by comparing Line 3 with Line 2 and Line 5 with Line 4, we observe that the inclusion of the DDF module yields a larger improvement when combined with Uni-SA (+0.50\% vs. +0.27\% mIoU), even though Line 4 starts at a higher level. This indicates that Uni-SA facilitates better fusion between dense satellite features and sparse street features, which is consistent with the qualitative comparison shown in Figure \ref{fig: Qualitative_feature}.

\noindent\textbf{Different Fusion Strategies:} We conduct an ablation study on various fusion strategies across satellite and street-view BEV features, as shown in Table \ref{tab:ablation_fusion}. The approach in Line 1 employs an attention mechanism for fusion \cite{gao2024complementing}, which undermines features of dynamic regions, limiting its suitability to static 2D mapping and making it unsuitable for dynamic 3D occupancy prediction. Lines 2 and 3 employ the fundamental fusion strategies of addition (add) and concatenation (concat), respectively. These do not substantially disrupt the dynamic features of the street branch but introduce erroneous features from the satellite in dynamic areas. Our method (Line 5) achieves the best performance due to the decoupling of dynamic and static regions. Additionally, alignment operations \cite{ye2024sg} are also ineffective in our experiments, possibly due to the complexity of the task or the lack of explicit supervision, as exhibited by the comparison between Lines 4 and 5.

\noindent\textbf{Impact of Satellite View Branch Components:} We conduct experiments on each component of the satellite-view branch, as shown in Table \ref{tab:ablation_satellite}. Compared to baseline \cite{yu2023flashocc}, the addition of the main structure of our satellite branch alone, namely the U-shaped feature extractor, achieves a +4.52\% mIoU improvement, with a dynamic gain of +4.1\% mIoU and static gain of +5.2\% mIoU. 
Further incorporating our 3D-Proj Guidance, which integrates static semantic segmentation and height estimation as auxiliary tasks into the satellite branch, yields an additional +0.22\% improvement in mIoU, with a static-specific gain of +0.91\% mIoU. This underscores the effectiveness of 3D-Proj Guidance in uncovering more comprehensive insights for static regions. Finally, adding the soft gated convolution yields a further +0.07\% mIoU improvement. This indicates the effectiveness of the module via high-frequency noise filtering. 

\begin{figure}[!t]
    \centering
    \includegraphics[width=\linewidth]{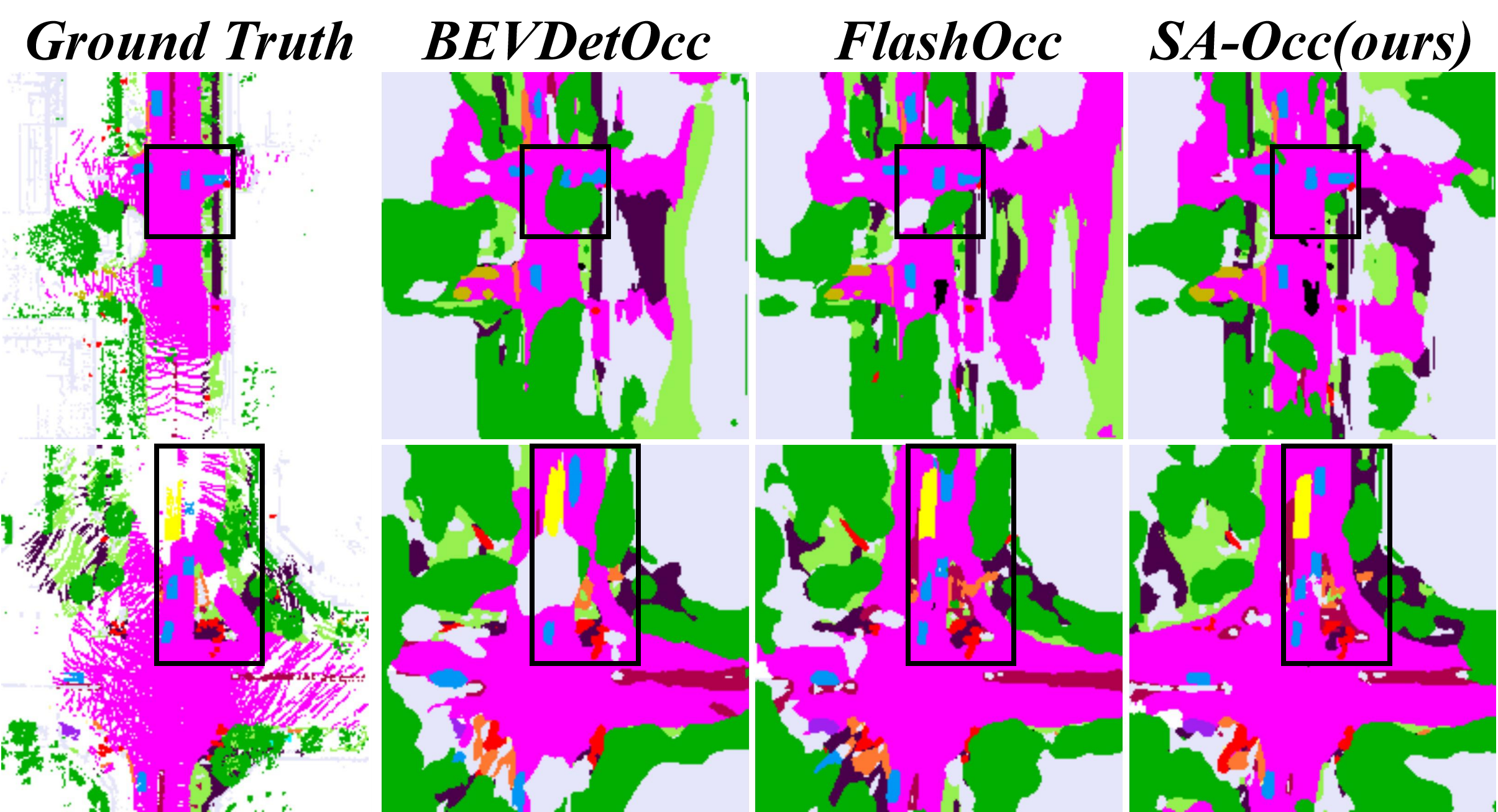}
    \caption{Qualitative Comparison of SA-Occ with Baselines. SA-Occ shows clearer boundaries and stronger robustness for both dynamic targets and static regions, especially in occluded areas.}
    \label{fig: qualitative-comparison}
\end{figure}

\noindent\textbf{Impact of DDF Components:} We also conduct an ablation study on the components of the Dynamic-Decoupling Fusion (DDF) module, as shown in Table \ref{tab:ablation_ddf}. Compared to Line 1, the Dynamic-Decoupling strategy (DDS) alone results in a +0.33\% mIoU improvement, with a dynamic gain of +0.65\% mIoU. This confirms that the DDS effectively mitigates the interference caused by asynchronous imaging of satellite data in dynamic regions. 
Further incorporating the Dynamic-encoding Spatial Attention (DSA) yields an additional +0.17\% mIoU improvement (0.1\% mIOU higher than the origin spatial attention), with a dynamic gain of another +0.32\% mIoU. 
The slight performance drop in static categories is attributed to the weakened interaction between satellite imagery and potential static objects along the z-axis within the dynamic BEV regions in 3D space. Despite this minor decrease, the overall mIoU remains largely unaffected, thanks to the significant gains of dynamic objects.

\noindent\textbf{Different Dynamic Area Label Source:} 
For the source of dynamic region annotations, we conduct an ablation study, as shown in Table \ref{tab:ablation_labels}. 
Experimental results show that labels derived from 3D detection bounding boxes outperform those from 3D occupancy voxel labels (+ 0.12\% mIOU). Furthermore, the union of both sources yields a further +0.13\% mIoU improvement. This indicates that more comprehensive dynamic region annotations enhance robustness by capturing both instance-level semantics (from bounding boxes) and fine-grained details (from voxel labels).  

\noindent\textbf{Balance between accuracy and speed:} We compared SA-Occ with our baseline \cite{yu2023flashocc} and its predecessor \cite{huang2021bevdet} as shown in Table \ref{tab:model_complexity}. Despite its higher parameter counts and FLOPs, SA-Occ achieves a 21.73\% accuracy improvement (6.97\% mIOU) with only a 6.93 ms increase in latency. Thanks to our parallelized architecture, SA-Occ still maintains a latency advantage (41.30\% lower than that of BEVDetOcc \cite{huang2021bevdet}). A qualitative comparison shown in Figure \ref{fig: qualitative-comparison} further highlights our substantial improvement.


\subsection{Comparisons with State-of-the-art}

As is shown in Table \ref{tab: 4}, we categorize state-of-the-art models into four groups based on their configurations: 1) single-frame models, 2) dual-frame models with ResNet50 \cite{he2016identity}, 3) multi-frame models, and 4) dual-frame models with SwinTransformer-Base \cite{liu2021swin} and a resolution of 512$\times$1,408.  

Our SA-Occ (V1), using a single-frame input, significantly outperforms all models in the first group: +6.97\% mIoU over FlashOcc (M1) \cite{yu2023flashocc}, +7.41\% mIoU over BEVDetOcc (1f) \cite{huang2021bevdet}, and +10.52\% mIoU over CTF-Occ \cite{tian2024occ3d}. Notably, SA-Occ (V1) outperforms dual-frame models in the second group, with +3.04\% mIoU over BEVDetOcc (2f) and +1.21\% mIoU over FlashOcc (M2), while remaining competitive with 8-frame BEVDetOcc and 16-frame FBOcc \cite{li2023fb} in the third group.

In the second group, SA-Occ (V2) adopts the BEVStereo \cite{li2023bevstereo} paradigm to scale up to a dual-frame method as \cite{yu2023flashocc, huang2021bevdet}. It outperforms competitors by +4.55\% mIoU over BEVDetOcc (2f) \cite{huang2021bevdet}, +2.81\% mIoU over FlashOcc (M2) \cite{yu2023flashocc}, and +3.37\% mIoU over BEVDetOcc (2f) with a larger resolution (384$\times$704). Remarkably, it surpasses FBOcc \cite{li2023fb} by 1.54\% mIoU and BEVDetOcc (8f) by 1.39\% mIoU in the third group, achieving superior performance with a more lightweight configuration.

SA-Occ (V3) exhibits consistent improvements in the third group. Compared to BEVDetOcc (8f) with a larger input resolution of 384$\times$704, it achieves a 2.43\% higher mIOU. It also achieves +2.58\% mIoU over FBOcc (E) \cite{li2023fb}, which uses 16 frames. Additionally, despite FastOcc \cite{hou2024fastocc} leveraging a heavier ResNet101 backbone, a larger input resolution of 640$\times$1,600, and 16 frames, SA-Occ (V3) still outperforms it by 2.48\% mIoU, and even +0.94\% mIoU over its test-time augmented version.

In the last group, SA-Occ (V4) follows the configuration of \cite{huang2021bevdet, yu2023flashocc} but without corresponding 
detection-task pretraining. It still outperforms FlashOcc (M3) by 0.48\% mIoU and BEVDetOcc \cite{huang2021bevdet} by 1.88\% mIoU. Upgrading the satellite branch backbone to ResNet50 in SA-Occ (V5) improves performance by 0.61\% mIoU over V4. Extending training to 36 epochs further enhances performance by 0.35\% mIoU, surpassing BEVDetOcc \cite{huang2021bevdet} by 2.19\% mIoU.

\section{Conclusion}
\label{sec:conclusion}
We present SA-Occ, the first satellite-assisted 3D occupancy prediction framework, accompanied by a novel GPS/IMU-aligned satellite-street dataset that enables real-world application. We address key challenges in cross-view perception including: \textbf{1)} A Dynamic-Decoupling Fusion module to resolve temporal inconsistencies caused by non-real-time satellite imagery; \textbf{2)} A 3D-Proj-Guided Satellite Branch for robust satellite BEV features; and \textbf{3)} A Uniform Sampling Alignment module for bridging the sampling density gap. SA-Occ achieves a 6.97\% mIOU improvement at the cost of a 6.93 ms latency increase compared to FlashOcc. This work demonstrates the potential of satellite view to enhance 3D scene understanding, paving the way for future research in cross-view perception.
{
    \small
    \bibliographystyle{ieeenat_fullname}
    \bibliography{main}
}

\end{document}